\documentclass[times, twoside]{IEEEtran}
\usepackage{graphicx} 
\usepackage{caption}
\usepackage{amsmath}
\usepackage{makecell}
\usepackage{multirow}
\usepackage{hhline}
\usepackage[table]{xcolor}
\usepackage{pifont}
\usepackage{makecell}
\usepackage{float}
\usepackage{booktabs}

\usepackage{caption}       

\usepackage{array}
\usepackage{graphicx}
\usepackage{multirow}
 \usepackage{soul}

\usepackage{authblk}
\date{}

\usepackage[table]{xcolor}
\usepackage[margin=1in]{geometry}
\usepackage{tabularx}

\renewcommand{\tablename}{Table}
\renewcommand{\thetable}{\arabic{table}}

\author[1]{Shubham Agarwal}
\author[1]{Thomas Searle}
\author[1]{Mart Ratas}
\author[2]{Anthony Shek}
\author[2,3]{James Teo}
\author[1,4]{Richard Dobson}
\affil[1]{Department of Biostatistics \& Health Informatics, King's College London, London, U.K.}
\affil[2]{Guy's and St Thomas’ NHS Foundation Trust, London, UK}
\affil[3]{Department of Neurology, King's College Hospital NHS Foundation Trust, London, UK}
\affil[4]{Health Data Research UK London and Institute of Health Informatics, University College London, London, UK}

\affil[ ]{\textit {shubham.agarwal@kcl.ac.uk}}

\begin{document} 
\title{Improving Extraction of Clinical Event Contextual Properties from Electronic Health Records: A Comparative Study}

\maketitle

\begin{abstract}
Electronic Health Records are large repositories of valuable clinical data, with a significant portion stored in unstructured text format.
This textual data includes clinical events (e.g., 
 disorders, symptoms, findings, medications and procedures) in \textit{context} that if extracted accurately at scale can unlock valuable downstream applications such as disease prediction. 
Using an existing Named Entity Recognition and Linking methodology, MedCAT, these identified concepts need to be further classified (contextualised) for their relevance to the patient, and their temporal and negated status for example, to be useful downstream. 
 \newline
This study performs a comparative analysis of various natural language models for medical text classification. 
Extensive experimentation reveals the effectiveness of transformer-based language models, particularly BERT. When combined with class imbalance mitigation techniques, BERT outperforms Bi-LSTM models by up to 28\% and the baseline BERT model by up to 16\% for recall of the minority classes. The method has been implemented as part of CogStack/MedCAT framework and made available to the community for further research.

\end {abstract}

\section{Introduction}
\label{sec:introduction}
Electronic Health Records (EHRs) contain a record of interactions with the patient and all the data concerning health that has been made by, or on behalf of, a health professional. This can be in connection with the diagnosis, care or treatment of the individual or a \textit{secondary} use that is non-clinical, administrative or research purpose \cite{nhs_ehr}.
This information is contained in multiple formats, with unstructured text being a significant proportion \cite{hayrinen2008definition}. Clinical text classification is a vital step in the sequence of tasks that facilitate the extraction of clinical information. These tasks can unlock tremendous opportunities for large-scale systemic analysis \cite{spasic2020clinical}, spanning from the detection and prediction of adverse events \cite{tayefi2021challenges}, to cancer pathology report coding \cite{tayefi2021challenges}, and improving the care quality \cite{menachemi2011benefits} among numerous others.

Before text classification, we perform a named entity recognition and then linking task (NER+L) to extract the raw text to clinical events such as a diagnosis, symptom, finding or procedure, and link each span to a standardised clinical terminology. For example, in the text "patient has been confirmed a diagnosis of diabetes", the NER+L task will extract the entity `diabetes' as the diagnosis `diabetes mellitus' and link for example the SNOMED CT \cite{snomed} identifier: SCTID: 73211009. We use the MedCAT library for NER+L, an openly available, and easily fine-tunable method \cite{kraljevic2021multi}. \\
After NER+L, further contextualization is often required to ensure that the extracted entities capture the context of how the entity appears in the text. This can be referred to as an entity attribute \cite{savova2010mayo}, property, modifier, or a meta-annotation in the MedCAT context. 
The meta-annotation categories we consider in this work are:
\begin{itemize}
    \item Presence (Not present $\vert$ Hypothetical $\vert$ Present) - to determine if the entity is negated, positively or hypothetically mentioned. 
    \item Experiencer (Other $\vert$ Family $\vert$ Patient) - to determine if the entity was experienced by the patient, family member or is referred to in some other way.
    \item Temporality (Past $\vert$ Future $\vert$ Recent) - to determine the time of the medical entity
\end{itemize}

\begin{figure} [ht]
    \centering
    \includegraphics[width=1\linewidth]{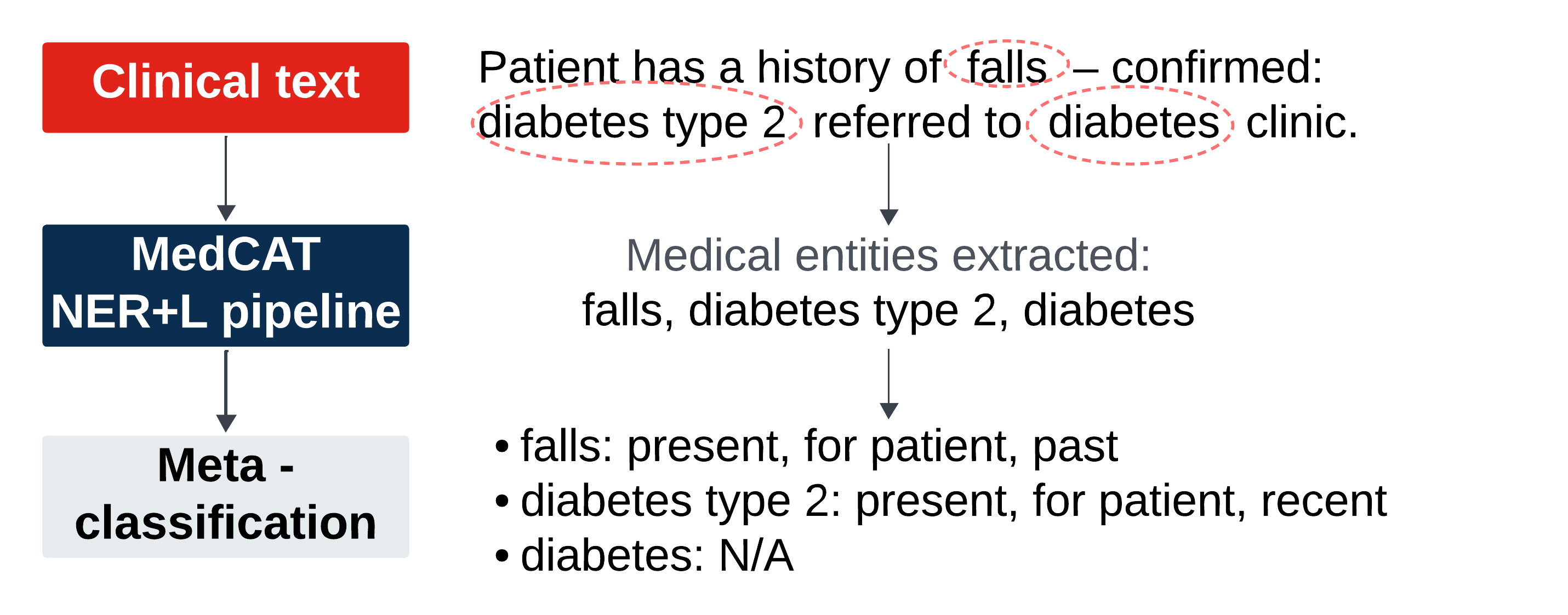}
    \caption{Example output for context meta-annotation}
    \label{fig:fig1-eg}
\end{figure}

\noindent Fig \ref{fig:fig1-eg} describes an example clinical text and meta-classification output for it.

Text classification, particularly in the medical domain, is challenging due to the complexity of the data, the extensive use of medical jargon, the sensitive nature of the information, and the presence of inconsistent or missing data \cite{ratwani2017electronic}.
Additionally, medical data often suffers from class imbalance, which presents further challenges for applied algorithms \cite{khushi2021comparative}. \newline
To address these challenges, researchers have explored the use of Bi-directional Long-Short Term Memory (Bi-LSTM) \cite{mascio2020comparative} and transformer models, specifically Bidirectional Encoder Representations (BERT) models \cite{si2019enhancing} \cite{li2024artificial}. 

In this study, we conduct a comparative analysis of the performance of Bi-LSTM and BERT models for text classification on EHR data. We evaluate the impact and comprehensively assess the influence of class imbalance on model performance with an exploration of mitigation techniques to address the same. We utilise LLMs to generate synthetic data and explore the in-context learning performance for these tasks.

\section{Methodology}
\label{sec:methodology}

\subsection{Dataset Description} The dataset is sourced from CogStack \cite{jackson2018cogstack}, deployed at Guy's \& St Thomas' NHS Foundation Trust, and includes 1800 documents. This data contains patient records (EHR data) that are annotated using MedCAT \cite{kraljevic2021multi}.
The data is across multiple clinical specialities: geriatrics, nephrology, ENT and metabolic disorders. The annotated data contains information for the classification tasks with the data distribution mentioned in Table \ref{tab:datasets}.  

\begin{table} [ht] 

\caption{Dataset description}
\label{tab:datasets}
\begin{centering}
\begin{tabularx}{1\columnwidth}[t]{c c c}
\toprule
   \textbf{Category}   & \textbf{Class}   & \textbf{Number of samples}   \\ 
    \toprule
 \multirow{3}{*}{Presence}  & Not present (False) & 578  \\ 
  & Hypothetical (N/A) &  978 \\
   &  Present (True) & 7430  \\
\midrule
 \multirow{3}{*}{Experiencer} & Other & 1002  \\
 & Family & 75 \\
  & Patient & 7908  \\
\midrule
 \multirow{3}{*}{Temporality} & Past & 733  \\
& Future & 484  \\
  & Recent & 7771  \\
\bottomrule
\end{tabularx}
\end{centering}
\end{table} 

\subsection{Modelling BERT}

BERT, based on a multi-layer bidirectional Transformer architecture \cite{devlin2018bert}, leverages pre-training to achieve impressive results across various Natural Language Understanding (NLU) tasks. During pre-training, BERT is trained on massive amounts of plain text for masked language prediction and next sentence prediction tasks. This allows it to gain a deep understanding of language structure and relationships between words, even without labeled data. As a result, BERT often outperforms other models, especially when dealing with limited labeled data \cite{sun2019fine} \cite{khadhraoui2022survey}.

In this study, we use a BERT model to perform the described medical text classification task. The pre-trained BERT model (bert-base-uncased)\footnote{https://huggingface.co/google-bert/bert-base-uncased} is used with two fully connected layers added on top to perform the classification. Using two fully connected layers showed best performance on experimentation, improving the model’s ability to perform accurate and robust classification.

The medical entity being classified can be single or multiple words, and after tokenization, the single word can be broken down into multiple tokens.
BERT's sequence classifier output can be utilized, however, it is more pertinent to have the classification performed with respect to the medical entity, and not the entire sequence. This approach provides a more granular and detailed understanding of the text, and by classifying entities individually, the results become more targeted to the given entity rather than a general label for the whole sequence. Hence, the hidden states of the medical entity are retrieved from BERT's output and then passed through a max pooling layer.

\begin{figure*} [ht]
\centering
\includegraphics[width=1\textwidth]{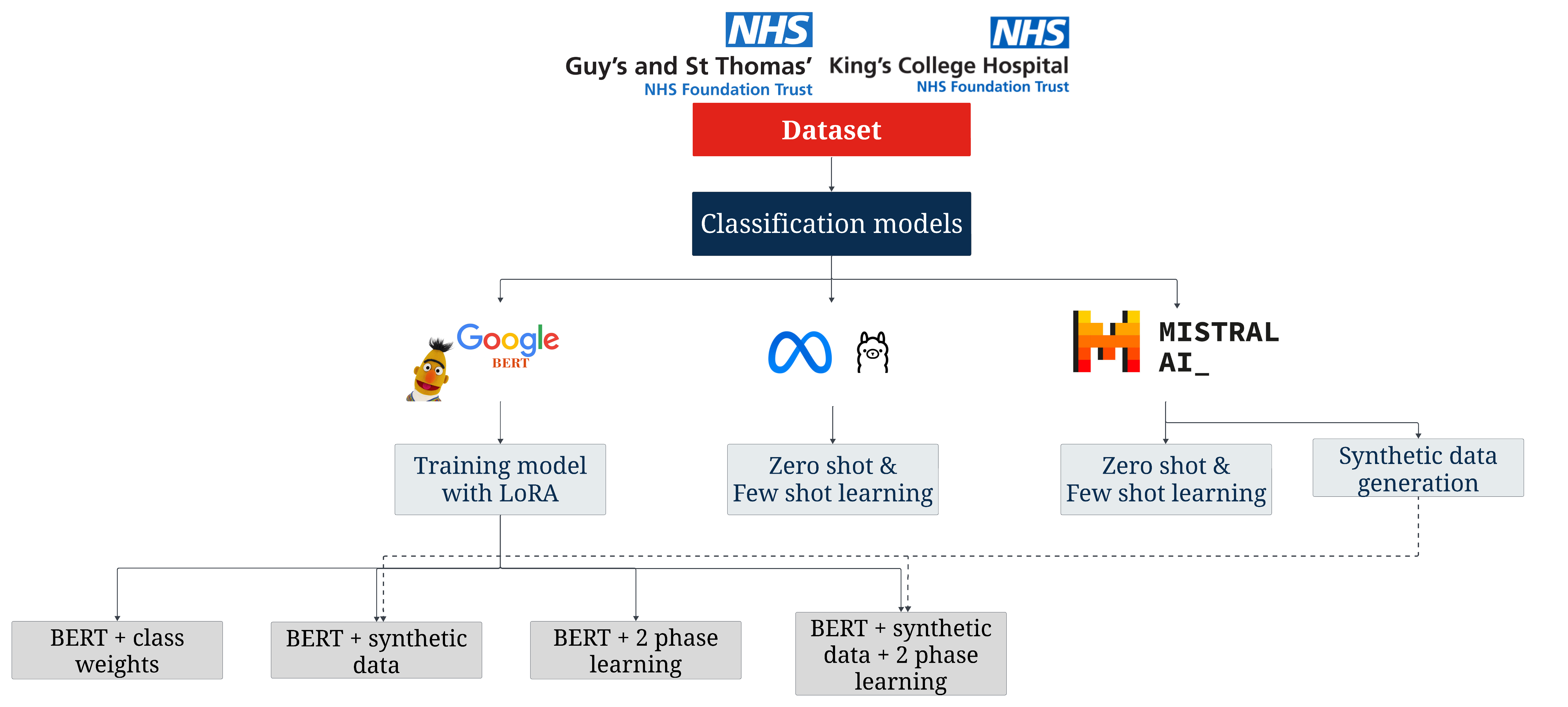}
\caption{Overview of modelling workflow}
\label{fig:computerNo}
\end{figure*}

From early experimentation, incorporating the representation of the entire sequence along with the medical entity improved performance over just including the embedding representing the medical entity. These vectors are then passed on to the fully connected layers. 
\newline Dropout of 0.2, AdamW optimizer function with learning rate scheduler, and batch size of 128 are used for training. Additionally, stratified splitting is employed to ensure that all classes are adequately represented in both the training and test datasets.

In this study, we experiment with BERT's parameters kept frozen and fine tuning BERT with LoRA \cite{hu2021lora}. Our experiments show that freezing BERT's parameters hinders its ability to learn complex data patterns, leading to inferior performance. LoRA-based fine-tuning overcomes this limitation, enabling effective model adaptation. This model is ablated with methodologies described in Section \ref{sec:class_imbal}.

\subsection{Class imbalance}
\label{sec:class_imbal}
Class imbalance occurs in supervised learning when a dataset has a significant skew in the number of examples across different classes. This means one or more classes (the majority classes) have far more data points than others (the minority classes). 
Class imbalance is a common challenge in real-world datasets, but it is particularly severe in clinical data \cite{kumar2022addressing}. Our dataset exemplifies this, as for Experiencer category, the 'Family' class has only 1\% of the data compared to the 'Patient' class (100x imbalance). Our efforts to collect additional data for underrepresented classes did not alter the class distribution, reflecting the real-world scenario where minority classes are much less common than the majority class, highlighting the inherent class imbalance challenge in our clinical data. 

Class imbalance in training data creates a challenge for training predictive models;  since the model encounters the majority class much more frequently during training, it tends to create a bias and prioritize learning the frequent class leading to the model performing poorly for the minority class \cite{li2010learning}.

To address the challenges presented by class imbalance, we use the following methodologies:

\begin{itemize}
    \item \textbf{Class weights}: Class weights can address class imbalance by giving different weights (importance) to the majority and minority classes. The difference in class weights impacts training by assigning higher weights to the minority class to penalize its misclassification while reducing the weight for the majority class encourages the model to learn and better recognize the minority classes \cite{johnson2019survey}.

    \item \textbf{Synthetic data generation using LLM}: Class imbalance presents challenges due to the difference in the amount of data for each class. One effective solution to this problem is to generate additional data for the underrepresented classes. Multiple studies have shown positive outcomes on model performance with oversampled data \cite{johnson2019survey}.
    Based on our experiments, we opted to use the Mistral 7b model \cite{jiang2023mistral} for data generation as it demonstrated superior data generation capabilities compared to Llama 3 8b \cite{dubey2024llama}. We generate additional data samples for the minority and majority classes, with a greater emphasis on the minority class. By generating data for all classes, we ensure a uniform enhancement across the dataset, mitigating biases and avoiding altering the distribution of any single class. 
    The model was presented with 10 examples from the dataset, 8 from the minority classes and 2 from the majority classes. Manual validation for quality control was performed to ensure integrity of the data. The generated synthetic data comprises less than 5\% of the total dataset, which prevents the data distribution from being significantly altered.

    \item \textbf{2 phase learning}: 2 phase learning is a training approach designed to fix the issue of the gradients being dominated by the majority class \cite{lee2016plankton}.
        The 2 phases in this approach are:
        \begin{itemize} 
            \item [$\circ$] Phase 1: In this phase, all classes are down sampled to a specified value \textit{N} (that is close to the number of samples for the minority class) and training is performed with higher class weights weights given to minority classes. Phase 1 allows the model to capture and learn the details for the minority classes.
            \item [$\circ$] Phase 2: During this stage, the model undergoes a second round of training, now on the entire dataset. The class weights assigned to minority classes are high but lower compared to the initial phase. This phase allows the model to capture the finer details for all classes, leading to a more finely-tuned model.
        \end{itemize}
    
\end{itemize}

\subsection{Large Language Models for classification}
Large Language Models (LLMs) have seen widespread usage in NLP and specifically in text classification tasks \cite{spasic2020clinical}. The latest state-of-the-art LLMs, like Llama 3 8b \cite{dubey2024llama} and Mistral 7b \cite{jiang2023mistral} represent the latest general purpose auto-regressive, large language models. These models have been pre-trained on large volumes of web scale data \cite{brown2020language}, then further pre-trained to follow instructions \cite{brown2020language}.  

To perform the classification, we leverage two techniques: \textbf{zero-shot learning} and \textbf{few-shot learning}. Zero-shot learning \cite{radford2019language} \cite{larochelle2008zero} is where model performs classification based only on the instructions in the prompt without any `training' examples \cite{rohrbach2011evaluating}. In few-shot learning \cite{wang2020generalizing}, the model is prompted with a limited set of examples (inputs and their corresponding outputs) alongside the classification instructions, enabling it to better understand the task at hand.
For few-shot learning, the models were provided with a total of 9 examples, distributed as 3 examples per class. The choice of 9 examples per task aims to maintain simplicity, clarity, and conciseness in the prompts, with longer prompts having the potential to reduce the model's effectiveness in performing these tasks \cite{brown2020language} \cite{sahoo2024systematic}.

\section{Results}
This section assesses the performance of the classification models using accuracy and F1-score. Due to the severe class imbalance in the dataset, macro F1-score is used to account for the performance of minority classes. Additionally, we use recall to provide a more comprehensive view of model performance across all classes.

\subsection{Performance of models} 
As shown in Tables \ref{tab:res_presence}, \ref{tab:res_exp}, and \ref{tab:res_tempo}, BERT models achieved higher macro F1-score and recall values for minority classes compared to the Bi-LSTM models. We delve deeper into the impact of the class imbalance mitigation techniques employed with Bi-LSTM and BERT in the following sections.

\subsection{Impact of synthetic data} To address class imbalance, we investigate the impact of oversampling data using LLM (Mistral). The models were trained on the original data supplemented with the synthetic data. The results reveal an overall performance improvement compared to the baseline models (both Bi-LSTM and BERT), with some exceptions. \\ For the Bi-LSTM model, oversampling demonstrably benefited \textbf{at least one minority class} in Presence, Experiencer and Temporality classification tasks.
For the BERT model, use of synthetic data show benefits for \textbf{at least one minority class} in both Presence and Experiencer classification. However, for the more intricate Temporality classification task, minimal improvement was observed.\\
This approach effectively aligned with our goal of enhancing performance for minority classes across both models.

\subsection{Impact of 2-phase learning} This section explores the impact of 2-phase learning on model performance. The results show improvements across all metrics for both models. 
For the Bi-LSTM model, 2-phase learning benefited \textbf{both minority classes} for the Experiencer and Temporality tasks, with a maximum of \textbf{9\%} and \textbf{5\%} increase in recall.
For the BERT model, the Presence and Temporality tasks yielded the most benefits. We observed a substantial performance improvement in \textbf{both minority classes}, with recall values increasing by up to \textbf{6\%} for Presence and \textbf{9\%} for Temporality. Impact for Experiencer task is more subtle, and while the approach did not greatly improve minority classes, it enhances performance for the majority class.

\subsection{Impact of 2-phase learning and synthetic data} This section explores the effect of combining 2-phase learning with synthetic data. This combined approach yields the best performing models overall (both Bi-LSTM and BERT), demonstrating performance boost across all metrics, particularly recall. 
\\For Bi-LSTM model, the performance improvements led to one minority class achieving a recall value above 0.8.

\begin{itemize}
    \item Presence and Temporality tasks exhibited a maximum of  \textbf{9\%} and \textbf{7\% improvements} in recall for  minority class.
    \item Experiencer classification achieved higher improvements, with maximum of \textbf{14\% increase} in recall for minority class.
\end{itemize}
For BERT model, the improvement in recall values for minority classes is noteworthy.
\begin{itemize}
    \item Presence and Temporality tasks exhibited a range of \textbf{5\% to 12\% improvements} in recall for both minority classes.
    \item Experiencer classification achieved the most gains, with \textbf{7\% and 16\% improvements} in recall for both minority classes.
\end{itemize}
Furthermore, BERT models consistently exhibited a \textbf{6\% improvement} in recall for the majority class in both Experiencer and Temporality tasks.

\subsection{Performance of LLMs for in-context classification} This section examines the performance of Llama and Mistral models in the context of few-shot learning for our classification tasks. As zero-shot learning yielded subpar results for both models, we are reporting on the few-shot learning outcomes. We plan to report on the enhanced performance later, once we apply the techniques discussed in section \ref{limitatons_and_future_work}. 
Both Llama and Mistral models exhibited limitations, particularly in recall values for minority classes. The lowest recall value observed was 0.05 for the Experiencer category (achieved by Llama). However, both models performed well on the majority class, with Llama reaching a high recall value of 0.97 for the Presence task.
Mistral achieved the best overall performance for LLMs for the Temporality task across all classes. Here, Mistral demonstrated recall values of 0.27 and 0.55 for its minority classes, and 0.74 for the majority class.
While the few-shot approach offers advantages, it did not yield optimal results, with further analysis performed in section \ref{sec:discussion}.

\setlength\tabcolsep{1.5pt}
\begin{table}[ht!]
\centering
\caption*{\centering{\textbf{Tables: Results for all classification tasks.} \newline CW - class weights in favour of minority classes \newline 2PL - 2 phase learning fine-tuning approach + CW, \newline SD - inclusion of synthetically generated data + CW}
\newline \textbf{Note:} In all tables, the first two are minority classes, and the third is the majority class.}

\caption{\centering{Model performance for Presence task}}

\label{tab:res_presence}

\begin{tabular}{ c c c c c c} 

 \toprule

 \multirow{2}{*}{ \textbf{Model}} &  \multirow{2}{*}{ \textbf{Accuracy}}&\textbf{Macro} &\multicolumn{3}{c}{\textbf{Recall}}  \\ 
 \cline{4-6}
  
 & &\textbf{F1-score} & \textbf{\textit{Not present}} &\textbf{\textit{N/A}} &\textbf{\textit{Present}}  \\
\toprule

 Bi-LSTM (w/ CW) & \textbf{0.89} & 0.78 & 0.77 & 0.72 & 0.93 \\
 Bi-LSTM (w/ SD) & 0.87 & 0.8 & 0.79 & 0.75 & 0.9 \\
 Bi-LSTM (w/ 2PL) & 0.88 & 0.81 & 0.76 & 0.77 & 0.91 \\
 \makecell{Bi-LSTM (w/ 2PL \\ + SD)} & \textbf{0.89} & 0.84 & 0.84 & 0.79 & 0.92 \\
 \midrule

BERT (w/ CW) & 0.86 & 0.82 & 0.8 & 0.77 & 0.91 \\

BERT (w/ SD) & 0.87 & 0.82 & 0.8 & 0.79 & 0.88 \\

BERT (w/ 2PL) & 0.88 & 0.85 & 0.85 & 0.78 & 0.91 \\

\textbf{\makecell{BERT (w/ 2PL \\ + SD)}} & \textbf{0.89} & \textbf{0.87} & \textbf{0.87} & \textbf{0.84} & 0.9 \\

\midrule

\makecell{Llama (few shot)} & 0.84 & 0.45 & 0.6 & 0.03 & \textbf{0.97} \\

\makecell{Mistral (few shot)} & 0.8 & 0.38 & 0.1 & 0.2 & 0.95 \\

 \bottomrule
\end{tabular}

\bigbreak

\centering
\caption{\centering{Model performance for Experiencer task}}
\label{tab:res_exp}
\begin{tabular}{ c c c c c c} 

 \toprule

 \multirow{2}{*}{ \textbf{Model}} & \multirow{2}{*}{ \textbf{Accuracy}}&\textbf{Macro} &\multicolumn{3}{c}{\textbf{Recall}}  \\ 
 \cline{4-6}
  
 & &\textbf{F1-score} & \textbf{\textit{Other}} &\textbf{\textit{Family}} &\textbf{\textit{Patient}}  \\
\toprule

 Bi-LSTM (w/ CW) & 0.9 & 0.77 & 0.77 & 0.64 & 0.92 \\

Bi-LSTM (w/ SD) & 0.91 & 0.78 & 0.75 & 0.68 & 0.92 \\
Bi-LSTM (w/ 2PL) & 0.92 & 0.82 & 0.83 & 0.7 & 0.93 \\

\makecell{Bi-LSTM (w/ 2PL \\ + SD)} & 0.92 & 0.83 & 0.84 & 0.73 & 0.93 \\

\midrule

BERT (w/ CW) & 0.87 & 0.84 & 0.83 & 0.81 & 0.9 \\

BERT (w/ SD) & 0.88 & 0.87 & 0.84 & 0.85 & 0.91 \\

BERT (w/ 2PL) & 0.91 & 0.87 & 0.82 & 0.82 & 0.94 \\

\textbf{\makecell{BERT (w/ 2PL \\ + SD)}} & \textbf{0.93} & \textbf{0.93} & \textbf{0.89} & \textbf{0.94} & \textbf{0.95} \\

\midrule

\makecell{Llama (few shot)} & 0.69 & 0.51 & 0.05 & 0.9 & 0.75 \\

\makecell{Mistral (few shot)} & 0.74 & 0.53 & 0.17 & 0.65 & 0.8 \\

 \bottomrule
\end{tabular}

\bigbreak 
\centering
\caption{\centering{Model performance for Temporality task}}
\label{tab:res_tempo}

\begin{tabular}{ c c c c c c} 

 \toprule

 \multirow{2}{*}{ \textbf{Model}} &  \multirow{2}{*}{ \textbf{Accuracy}}&\textbf{Macro} &\multicolumn{3}{c}{\textbf{Recall}}  \\ 
 \cline{4-6}
  
 & &\textbf{F1-score} & \textbf{\textit{Past}} &\textbf{\textit{Future}} &\textbf{\textit{Recent}}  \\
\toprule

 Bi-LSTM (w/ CW) & 0.87 & 0.79 & 0.72 & 0.78 & 0.91 \\
  
  Bi-LSTM (w/ SD) & 0.87 & 0.8 & 0.75 & 0.77 & 0.9 \\
  
  Bi-LSTM (w/ 2PL) & 0.87 & 0.81 & 0.74 & 0.82 & 0.91 \\

  \makecell{Bi-LSTM (w/ 2PL \\ + SYN)} & \textbf{0.91} & 0.84 & 0.75 & 0.84 & \textbf{0.93} \\
  \midrule

BERT (w/ CW) & 0.82 & 0.8 & 0.8 & 0.78 & 0.83 \\
BERT (w/ SD) & 0.84 & 0.81 & 0.79 & 0.79 & 0.85 \\

BERT (w/ 2PL) & 0.84 & 0.84 & 0.82 & 0.85 & 0.85 \\

\textbf{\makecell{BERT (w/ 2PL \\ + SD)}} & 0.87 & \textbf{0.86} & \textbf{0.84} & \textbf{0.86} & 0.89 \\
\midrule
\makecell{Llama (few shot)} & 0.8 & 0.43 & 0.1 & 0.36 & 0.9 \\

\makecell{Mistral (few shot)} & 0.77 & 0.47 & 0.27 & 0.55 & 0.74 \\

 \bottomrule
\end{tabular}
\end{table}

\section{Discussion}
\label{sec:discussion}
This study assessed the performance of Bi-LSTM and BERT models for medical text entity context classification across three tasks. We further investigated the effectiveness of various class imbalance mitigation techniques in enhancing performance, particularly for minority classes.

\subsection{Bi-LSTM and BERT models}
The results showcase BERT's superiority over the Bi-LSTM model, particularly in its ability to classify minority classes with high recall. Our experiments emphasise the importance of addressing class imbalance for robust model performance.
Bi-LSTM models coupled with class imbalance mitigation techniques, show \textbf{good performance for one minority class} - 'Not present' for Presence, 'Other' for Experiencer and 'Future' for Temporality task, but struggle to achieve high performance across both minority classes, whereas BERT models when combined with the class imbalance mitigation techniques, demonstrate \textbf{high performance across all classes} for all classification tasks. 
\\ In terms of computational efficiency, the Bi-LSTM model is lightweight and fastest among the models. In comparison, the BERT model is heavier and more computationally demanding, requiring, on average, 32\% more time to complete a single training epoch.

\subsection{Class imbalance mitigation techniques}
While using class weighting as a standalone technique does not yield optimal model performance, it serves as a crucial foundation for more sophisticated approaches to address class imbalance.

Generating synthetic data with LLMs represents a direct approach to class imbalance by generating synthetic data for minority classes. This strategy yielded notable improvements in recall for \textbf{at least one minority class} in each classification task (except Temporality task for BERT). With manual validation of generated data for quality control, oversampling shows promise to address class imbalance.

Our study also investigated 2-phase learning as a technique for mitigating class imbalance. It shows substantial improvements in \textbf{both minority classes for two} classification tasks, and noticeable improvement on majority class for the third task (only for BERT). This suggests that 2-phase learning can be effective in achieving a more balanced performance across classes. 
Its intuitive focus on minority classes initially, followed by fine-tuning on the complete dataset, aligns well with the goal of addressing class imbalance.

Finally, combining the two method: generating synthetic data and 2-phase learning had the most impact in addressing class imbalance. This approach achieved substantial gains,\textbf{ 5\% to 17\% in recall for minority} classes across all classification tasks for both models. Additionally, for one task, it led to a noticeable improvement of \textbf{6\% in majority} class recall. These findings suggest that a combination of these techniques can be effective in achieving balanced performance across all classes. Generating synthetic data with LLMs likely helps capture the characteristics of minority classes, while 2-phase learning ensures focus primarily on minority and then majority class during training. This combined approach presents a promising strategy to effectively manage class imbalance in text classification tasks.

\subsection{LLMs for in-context classification}
\label{llm_classification_analysis}
LLM for in-context classification exhibited limitations in consistently classifying minority classes with high recall, except for specific cases (e.g., Llama for the Experiencer task). Our investigation revealed that this is likely due to a bias towards the majority class. LLMs tend to classify a sample as the default class (majority class) unless there are clear and explicit indicators of the minority class. This approach struggles as the indicators for minority classes are often subtle and contextual, not always explicit. 
For example, for the Temporality task, LLMs classify most samples as "Recent" (majority class) unless explicitly mention of the past or future.
Example data sample: \textit{"The doctor intends to discuss corticosteroid treatment options if your asthma worsens".} It can be inferred from the example that the treatment is intended for future, however, it is not mentioned in explicit terms.

\subsection{Classification task analysis}
We analyzed the classification tasks to understand the complexity of each. The models performed best on the Experiencer task, which can be attributed to the distinct boundaries between the classes, enabling the models to more effectively capture the distribution. The Presence classification task is more complex, and while the 'Present' class is well-defined, there can be overlap between the 'Not present' and 'N/A' classes. Although they represent different concepts, the distinction between them in the dataset can be unclear. Lastly, the Temporality task is even more challenging.  The 'Recent' class is well-defined, however, the 'Future' and 'Past' classes exhibit variance. These classes can refer to time periods ranging from yesterday/tomorrow to x years. Additionally, the time periods are not always quantified, further complicating the data.

\subsection{Limitations and future work}
\label{limitatons_and_future_work}
The data for this study was sourced from a single NHS site (GSTT) which limits the generalisability of our findings to a certain degree. To address this, we plan to expand our dataset and run further experiments across multiple NHS sites that have CogStack deployments.
For the BERT model, we used the `\textit{bert-base}' variant; previous research has demonstrated that using the BERT large variant and domain-specific models like ClinicalBERT \cite{huang2019clinicalbert} and BioBERT \cite{lee2020biobert} can enhance performance. Future work will incorporate these models to further refine our results.
LLMs when used for in-context classification have a bias to majority class that hinders performance (discussed in depth in section \ref{llm_classification_analysis}), and to address this, we plan to tweak the prompts to encourage the inclusion of more subtle indicators of minority classes. We also aim to investigate the impact of using higher number of samples per class for few-shot prompting on performance and also utilize Human-in-the-loop and Chain-of-thought prompting techniques to boost performance \cite{wei2022chain}.

\section{Conclusion}
In this study, we analyzed the performance of Bi-LSTM and BERT models for contextual classification of entities extracted from clinical text by MedCAT's NER+L. Also, we investigated the effectiveness of class imbalance mitigation techniques. 

BERT models outperformed Bi-LSTM models, especially on recall for minority classes. 
Addressing class imbalance significantly impacted model performance. Using LLMs to generate synthetic data and 2-phase learning together yielded substantial improvements, particularly for minority classes. This suggests a powerful strategy for handling class imbalance in medical text classification.
We observed limitations in using LLMs for classification, as they have a bias towards the majority class. Future work may improve this performance, however usage of these models across large clinical corpora is expensive and cumbersome especially for this use case that can require millions of classifications for even a relatively small dataset. 

The research advances the field of medical NLP by demonstrating the effectiveness of BERT for extracting clinical event data from unstructured medical text through text classification. As part of MedCAT, an open-source tool, it facilitates collaboration and widespread adoption across various healthcare settings, potentially leading to substantial improvements in patient care.

\section*{Acknowledgments}
We appreciate the help and support from GSTT, the GSTT-Cogstack team and Aleksandra Foy for helping with data collection in this work.
This work was supported by Health Data Research
UK, an initiative funded by UK Research and
Innovation, Department of Health and Social Care
(England) and the devolved administrations, and
leading medical research charities. SA, TS, RD are part-funded by the National
Institute for Health Research (NIHR) Biomedical
Research Centre at South London and Maudsley
NHS Foundation Trust and King’s College
London. RD is also supported by The National Institute
for Health Research University College London
Hospitals Biomedical Research Centre.
This paper represents independent research
part funded by the National Institute for Health
Research (NIHR) Biomedical Research Centre at
South London and Maudsley NHS Foundation
Trust and King’s College London. The views expressed are those of the authors and not necessarily
those of the NHS, the NIHR or the Department of
Health and Social Care. The funders had no role in
study design, data collection and analysis, decision
to publish, or preparation of the manuscript.

\bibliographystyle{IEEEtran}  
\bibliography{Article}

\begin{appendix}

\section{LLM Prompts}

\subsection{Example of the LLM prompt used for zero and few shot}

"""
    [INST]You are a text classification bot.
    Your task is to assess intent and categorize the input
    text into one of the following predefined categories:
    2: Experiencer - Patient / default,
    1: Experiencer - Family,
    0: Not applicable
    \newline Explanation of labels:
    Label 2 (patient / default) is the class where the context strongly indicates that the given medical entity is for the patient. The text will not explicitly contain mention that it is for the patient, you have to infer it.
    Label 1 (family) is the class where the context clearly indicates that the given medical entity is for the family.
    Label 0 (not applicable) is when the input data does is not applicable to the category.
   \\  
   You will only respond with the predefined category. Do not provide explanations or notes.
    \newline Inquiry: {text}
    \newline Category: [/INST]
    """
\newline
\subsection{Examples Generated from LLMs}
\textit{\textbf{For Experiencer:}}
\begin{itemize}
    \item His younger sibling is receiving chemotherapy for colon cancer. They attend oncology visits together; 'colon cancer' - Family
    \item The physician diagnosed her with Hodgkin Lymphoma during last tuesday's session; 'Hodgkin Lymphoma' - Patient
    \item The support group aimed at creating awareness among individuals suffering from multiple sclerosis in their community; 'multiple sclerosis' - Other 
\end{itemize}
\vspace{1.5mm} \textit{\textbf{For Presence:}}
\begin{itemize}
    \item At my annual checkup, the GP recommended having a colonoscopy due to family history; 'colonoscopy' - Present
    \item Patients who have severe kidney damage might require dialysis therapy temporarily or permanently; 'kidney damage' - N/A
    \item Upon reviewing the patient's file, it appears there have been no diagnoses related to asthma or allergies; 'asthma' - Not present
\end{itemize}

\vspace{1.5mm} \textit{\textbf{For Temporality:}}
\begin{itemize}
    \item Based on current symptoms and test results, the patient will require hip replacement surgery in a couple of months; 'hip replacement surgery' - Future
    \item The patient underwent routine mammography today and has received the imaging results; 'mammography' - Recent
    \item Past X-ray examination indicated signs of osteoporosis, calling for medications and lifestyle changes; 'osteoporosis' - Past
\end{itemize}

\end{appendix}

\end{document}